# Computational analysis of the language of pain: a systematic review

**Diogo Afonso Pedro Nunes (Diogo A.P. Nunes) – corresponding author**

[1] Instituto de Engenharia de Sistemas e Computadores - Investigação e Desenvolvimento, Rua Alves Redol, 9, 1000-029 Lisboa, Portugal.

[2] Instituto Superior Técnico, Universidade de Lisboa, Avenida Rovisco Pais, 1049-001 Lisboa, Portugal.

diogo.p.nunes@inesc-id.pt

0000-0002-6614-8556

**Joana Maria de Pinho Ferreira Gomes (Joana Ferreira-Gomes)**

[3] Instituto de Investigação e Inovação em Saúde da Universidade do Porto (I3S). Rua Alfredo Allen 208, 4200-393 Porto, Portugal.

[4] Instituto de Biologia Molecular e Celular (IBMC), Universidade do Porto. Rua Alfredo Allen 208, 4200-393 Porto, Portugal.

[5] Departamento de Biomedicina – Unidade de Biologia Experimental, Faculdade de Medicina, Universidade do Porto. Alameda Prof. Hernâni Monteiro 4200-319 Porto, Portugal.

jogomes@med.up.pt

0000-0001-8498-6927

**Fani Lourença Moreira Neto (Fani Neto)**

[3] Instituto de Investigação e Inovação em Saúde da Universidade do Porto (I3S). Rua Alfredo Allen 208, 4200-393 Porto, Portugal.

[4] Instituto de Biologia Molecular e Celular (IBMC), Universidade do Porto. Rua Alfredo Allen 208, 4200-393 Porto, Portugal.

[5] Departamento de Biomedicina – Unidade de Biologia Experimental, Faculdade de Medicina, Universidade do Porto. Alameda Prof. Hernâni Monteiro 4200-319 Porto, Portugal.

fanineto@med.up.pt

0000-0002-2352-3336

**David Manuel Martins de Matos (David Martins de Matos)**

[1] Instituto de Engenharia de Sistemas e Computadores - Investigação e Desenvolvimento, Rua Alves Redol, 9, 1000-029 Lisboa, Portugal.

[2] Instituto Superior Técnico, Universidade de Lisboa, Avenida Rovisco Pais, 1049-001 Lisboa, Portugal.

david.matos@inesc-id.pt

0000-0001-8631-2870

**Word count of the main body of the manuscript:** 3997
**Supplementary materials:** 1
**Tables**: 4
**Figures**: 2

**ABSTRACT**

**Objectives:** This study aims to systematically review the literature on the computational processing of the language of pain, or pain narratives, whether generated by patients or physicians, identifying current trends and challenges.

**Methods:** Following the PRISMA guidelines, a comprehensive literature search was conducted to select relevant studies on the computational processing of the language of pain and answer pre-defined research questions. Data extraction and synthesis were performed to categorize selected studies according to their primary purpose and outcome, patient and pain population, textual data, computational methodology, and outcome targets.

**Results:** Physician-generated language of pain, specifically from clinical notes, was the most used data. Tasks included patient diagnosis and triaging, identification of pain mentions, treatment response prediction, biomedical entity extraction, correlation of linguistic features with clinical states, and lexico-semantic analysis of pain narratives. Only one study included previous linguistic knowledge on pain utterances in their experimental setup. Most studies targeted their outcomes for physicians, either directly as clinical tools or as indirect knowledge. The least targeted stage of clinical pain care was self-management, in which patients are most involved. Affective and sociocultural dimensions were the least studied domains. Only one study measured how physician performance on clinical tasks improved with the inclusion of the proposed algorithm.

**Discussion:** This review found that future research should focus on analyzing patient-generated language of pain, developing patient-centered resources for self-management and patient-empowerment, exploring affective and sociocultural aspects of pain, and measuring improvements in physician performance when aided by the proposed tools.

**Keywords:** pain; language of pain; natural language processing; systematic review; computational analysis

**LAY SUMMARY**

The objective of this study is to conduct a thorough literature review of the computational processing of pain language, or pain narratives, whether originating from patients or physicians, with the aim of identifying prevailing trends and challenges. An exhaustive, systematic literature review was undertaken to select pertinent studies that address predefined research questions. Selected studies were categorized according to their primary purpose and outcome, patient and pain population, textual data, computational methodologies, and outcome targets. Tasks performed by the selected studies included patient diagnosis and triaging, pain mention identification, treatment response prediction, biomedical entity extraction, correlation of linguistic features with clinical states, and lexico-semantic analyses of pain narratives. Most studies used physician-generated pain language from clinical notes. The results of this literature review suggest future research should concentrate on analyzing patient-generated pain language, developing patient-centric resources to facilitate self-management and improve patient-empowerment, exploring the affective and sociocultural facets of pain, and assessing improvement in physician performance when aided by the proposed tools.

# BACKGROUND AND SIGNIFICANCE

Pain is a subjective experience, embedded in the biopsychosocial model of health [1,2]. It is also private, unless communicated. There are many types of pain communication, such as facial expressions, verbal interjections, and narratives. Given the rich connection between language, psychology, and socio-culture, patient pain narratives reflect the multiple dimensions of pain in a more natural and personalized way. They can elucidate what it means to experience pain to that specific person and, when properly assessed, improve the clinical outcome [3], with information the person in pain believes to be relevant about the bodily distribution of the feeling of pain, temporal patterns of activity, intensity, emotional and psychological impacts, and others [4]. Indeed, patient pain utterances have been lexically [5] and grammatically [6,7] analyzed and related with clinical states, resulting namely in the widely used McGill Pain Questionnaire (MPQ) [8].

Patient pain narratives are an instance of the language of pain, which encompasses the linguistic formulation of the pain experience [7]. Anyone verbally expressing or describing an experience of pain (theirs or that of others) is effectively employing the language of pain. Although most practices and clinicians already include some kind of analysis of the natural language of pain, usually based on the clinician's own experience in communicating with people in pain and assessing them, they seldom follow a systematic, quantifiable, and explainable approach. Moreover, patients commonly report dissatisfaction in feeling heard or understood, and many clinicians find this communication challenging and frustrating [3].

Computational models assessing the language of pain may help supporting clinical pain assessment in a systematic, quantifiable, and explainable manner. Indeed, language has been explored with increasingly more complex natural language processing (NLP) techniques, due to the development of these techniques and the larger availability of computational resources and relevant data. In what respects health-related applications, various works started to focus on mental health topics, such as depression diagnosis [9], suicidal ideation detection [10], and the linguistic analysis of multiple and co-occurring mental health conditions [11], because they are closely related with language. Some focus has also been given to computationally explore the language of pain.

In this work we systematically review the literature on the computational processing of the language of pain, generated either by patients or physicians. To the best of our knowledge, at the time of writing, there is no published systematic review defining the landscape of this research paradigm.

## Research questions
This work aims at answering the following research questions:
- Q1. Given patient-generated and physician-generated language of pain:
    - Q1.a. Which language type is most analyzed?
    - Q1.b. Which tasks are performed for each language type?
- Q2. Are the linguistic findings by [5], [6], and [7] explicitly incorporated into the main processing?
- Q3. For whom are the studies and their outcomes primarily designed: patients (e.g., pain journaling app) or physicians (e.g., diagnosis assistant)?
- Q4. Which dimensions of pain are most and less studied?
- Q5. Which stage of pain care is being targeted: diagnosis, clinical-decision support, treatment/ rehabilitation, or self-management?
- Q6. Does physician performance on clinical tasks improve with the inclusion of the proposed algorithm?

# MATERIALS AND METHODS

We followed the PRISMA [12] guidelines. All steps were performed by at least two independent reviewers, and all disagreements were resolved through consensus, with the assistance of additional reviewers when necessary (up to four). Extraction was refined after piloting. Synthesis was validated by all reviewers.

## Article retrieval

This systematic review has two main axes: computational processing and language of pain. Table A.1 in Supplementary Materials shows the defined search keywords. The article search query was defined as the matching of at least one keyword in both axes, either in the publication title or abstract. We searched the following six databases: PubMed [1], IEEE Xplorer [2], ACM Digital Library [3], SCOPUS [4], Web of Science [5], and PLOS [6]. All databases were initially queried, without constraints, on January 20, 2023, except for SCOPUS, which was queried on January 23, 2023. We performed the same query one year later to update the retrieval for the full year of 2023.

## Article selection

We selected articles that met all the following criteria: (1) primary data source is the natural language of pain, (2) main processing is based on a computational method, and (3) primary outcome is some aspect of the experience of pain. We excluded articles that matched any of the following criteria: (1) is a duplicate of another article, (2) is not an original research article, (3) is not written in Portuguese, English, Spanish, or French, (4) abstract or full text are not available, or (5) does not meet all the inclusion criteria.

## Data extraction

We defined a set of data points primarily aimed at answering the review questions previously presented. These data points were based on [13] and [14], which performed systematic reviews on NLP in the clinical domain, and on [15] and [16], which performed systematic and scoping reviews on machine learning specifically in the pain domain. Table A.2 in Supplementary Materials shows the data points extracted from each selected article. We measured machine learning reporting quality using [17]. Studies were not excluded due to machine learning reporting quality.

# RESULTS

We retrieved a total of 2,394 articles. We performed study selection using Rayyan [7] [18]. A total of 40 articles were included for data extraction and synthesis. Fig. 1 shows the corresponding selection PRISMA flow diagram for the selection phase. Fig. 2 shows the distribution of studies per publication year.

## Primary purpose and task

Classification-based studies (n=17) performed patient diagnosis/ phenotyping (n=7), identification of pain mentions (n=3), treatment response prediction (n=2), identification of pain assessments (n=1), identification of patient-reported outcomes (PROs; n=1), patient questionnaire automation (n=1), risk factor prediction (n=1), and patient triaging (n=1).

---

[1] https://www.pubmed.ncbi.nlm.nih.gov
[2] https:// www.ieeexplore.ieee.org/Xplore/home.jsp
[3] https:// www.dl.acm.org
[4] https://www.scopus.com
[5] https://www.clarivate.com/webofsciencegroup/solutions/web-of-science
[6] https:// www.plos.org
[7] https://www.rayyan.ai

Correlation-based studies (n=5) determined the linguistic characteristics that correlated with specific clinical parameters, in this case, pain catastrophizing (n=3), pain intensity (n=3), illness intrusiveness (n=2), life satisfaction (n=1), depression (n=1), and treatment response (n=1).

Extraction-based studies (n=8) extracted pain intensity (n=3), body locations (n=3), drugs and treatments (n=3), pain qualities (n=2), pain care quality (PCQ) indicators (n=2), co-morbidities (n=2), symptoms (n=1), International Classification of Functioning, Disability, and Health (ICF) factors (n=1), reliefs (n=1), family environment (n=1), social determinants of health (SDoH; n=1), pain mentions (n=1), and Unified Medical Language System (UMLS) [19] semantic types (n=1).

Lexico-semantic analysis-based studies (n=10) determined the lexical and thematic characteristics of pain narratives, specifically reports of the overall experience of pain (n=5), of pain management (n=4), pain terms (n=1), and pain qualities (n=1).

Table 1 shows these data.

Table 1. Primary purpose and task.

| Ref. | Primary purpose | Primary task | Task extension |
|---|---|---|---|
| [20] | Develop a generalizable model to identify patients with chronic low back pain (LBP) from electronic health records (EHRs). | Classification | diagnosis/phenotype |
| [21] | Investigate clinical clues capable of distinguishing temporomandibular (TMD)-mimicking conditions from actual TMD. | Classification | diagnosis/phenotype |
| [22] | Identify acute LBP conditions from EHRs where it is not explicitly coded. | Classification | diagnosis/phenotype |
| [23] | Analyze the difference in frequency of back pain complaints on Twitter before and during the COVID-19 pandemic. | Classification | diagnosis/phenotype |
| [24] | Classify self-reported narratives of migraine or cluster headache patients. | Classification | diagnosis/phenotype |
| [25] | Validate use of language features from patient narratives for pain intensity estimation. | Classification | diagnosis/phenotype |
| [26] | Use natural language processing (NLP) to determine the severity of pain experienced by Osteoarthritis patients based on EHRs. | Classification | diagnosis/phenotype |
| [27] | Identify pain assessments from EHRs to assess pain care quality (PCQ) indicators. | Classification | identify pain assessment |
| [28] | Automate both the creation and utilization of regular expressions (RegEx) for the classification of clinical text. | Classification | identify pain mentions |
| [29] | Identify reports of pain severity in EHRs and classify positive notes according to the overall pain of the patient, with a generalizable model. | Classification | identify pain mentions |
| [30] | Identify "pain relevant" EHRs of sickle cell disease patients and classify "pain change" in positive cases. | Classification | identify pain mentions and changes |
| [31] | Classify symptom attributes (pain interference and fatigue domains) from textual meaning units relevant for patient-reported outcome (PRO) analysis from patient interviews. | Classification | identify PRO |
| [32] | Automate patient-reported outcome measures (PROMs) of knee injuries by developing an open-ended questionnaire that automatically maps onto Knee Injury and Osteoarthritis Outcome Score (KOOS) [33]. | Classification | questionnaire automation |
| [34] | Develop a novel deep learning approach to detect risk factors for underlying disease in patients presenting with LBP in EHRs. | Classification | risk factors |
| [35] | Classify placebo responders vs nonresponders in chronic back pain using language features, from patients undergoing treatment. | Classification | treatment response |
| [36] | Predict placebo and drug response in patients with chronic back pain before treatment, assessing model generalizability from [35]. | Classification | treatment response |
| [37] | Triage patients with musculoskeletal conditions based on primary care referral letters. | Classification | triage |
| [38] | Validate association between reattribution of chronic back pain to mind/brain processes and pain reduction in pain reprocessing therapy. | Correlation | pain catastrophizing, illness intrusiveness, pain intensity |
| [39] | Create a longitudinal pattern study of patient pain from EHRs. | Correlation | pain intensity |
| [40] | Classify and correlate risk for opioid agreement violation for chronic noncancer pain patients from EHR data (structured and unstructured), prior to the agreement. | Correlation | treatment response |
| [41] | Find correlations between language use and pain catastrophizing in chronic musculoskeletal pain patients. | Correlation | pain catastrophizing |
| [42] | Compare the performance of automatic feature extraction versus human feature extraction to predict measures of psychological and physical health for chronic pain patients. | Correlation | pain catastrophizing, life satisfaction, illness intrusiveness, pain intensity, depression |
| [43] | Extract information about various pain characteristics in EHRs to learn relations between pain symptoms, their identifying parameters, and treatments. | Extraction | body location, quality of pain, quantity of pain, symptoms, drugs/treatment |
| [44] | Map phrases from EHRs to classes in a pre-defined ontology (in this case, a chronic pain ontology). | Extraction | body location, quantity of pain, drugs/treatments, co-morbidities, reliefs, family environment |
| [45] | Extract International Classification of Functioning, Disability, and Health (ICF) factors from EHRs of LBP patients, relevant to the timing of rehabilitation. | Extraction | ICF factors |
| [46] | Develop a corpus of pain annotations from mental health records for use in downstream NLP pipelines. | Extraction | pain mention (and negation), body location, quality of pain, treatment |
| [47] | Extract PCQ indicators from EHRs. | Extraction | PCQ indicators |
| [48] | Identify PCQ indicators and assess patterns across different clinic visit types using NLP on Veteran's Health Administration (VHA) chiropractic clinic documentation. | Extraction | PCQ indicators |
| [49] | Apply NLP and inference methods to extract social determinants of health (SDoH) from EHRs of patients with chronic LPB. | Extraction | SDoH, depression, anxiety, quantity of pain |
| [50] | Triage patients with musculoskeletal conditions based on general practitioner referral letters. | Extraction | UMLS semantic types |
| [51] | Explore language use and themes of fibromyalgia patients on an online forum and compare with clinical knowledge. | Lexico-semantic analysis | pain experience |
| [52] | Understand what people in Ireland discuss on Twitter regarding their experience of chronic pain, their sentiment, gender, and tweet dissemination. | Lexico-semantic analysis | pain experience |
| [53] | Explore what patients discuss on social media regarding their chronic pain. | Lexico-semantic analysis | pain experience |
| [54] | Understand how the experience of pain is discussed online for sufferers of endometriosis. | Lexico-semantic analysis | pain experience |
| [55] | Understand how the experience of musculoskeletal disorder is communicated between patients, general practitioners, and specialists. | Lexico-semantic analysis | pain experience, pain management |
| [56] | Find differences in the themes discussed by nurses (EHRs) vs. patients (social media) when describing experiences of pain. | Lexico-semantic analysis | pain management |
| [57] | Explore discussed themes of opioid use for pain management on social media, to learn patient perspective that may go beyond the scope of standard questionnaires. | Lexico-semantic analysis | pain management |
| [58] | Thematic analysis of conversations between physicians and patients about chronic pain management. | Lexico-semantic analysis | pain management |
| [59] | Develop and make public a lexicon for "pain" based on multiple sources, for downstream NLP tasks. | Lexico-semantic analysis | pain terms |
| [60] | Update McGill Pain Questionnaire (MPQ) [8] vocabulary according to pain descriptor usage on social media. | Lexico-semantic analysis | quality of pain |

## Patient and pain population

Not all studies reported patient population distributions. Most reported the number of patients included in their work (n=28), ranging from 5 to 127,871 patients, with an average of 6,386.2 (Q2=227, Q3=713.5). The average reported patient mean age (n=14) was 46.1 (Q2=48.8, Q3=53.7). The average reported percentage of female-identified gender patients (n=17) was 57.1% (Q2=55.3%, Q3=72.4%).

Similarly, not all studies reported pain population distributions. Studies specifying the target chronicity of pain were specifically designed for chronic pain (n=19), acute pain (n=2), or both (n=4). Musculoskeletal pain (n=21) was the most targeted pain, followed by cancer (n=3), noncancer (n=2), nonterminal (n=1), and headache (n=1). Lower back and back were the most targeted pain locations (n=10).

Table 2 shows these data.

Table 2. Patient and pain population.

| Ref. | Number of patients | Average age | Percentage female | Chronicity of pain | Generic type of pain | Specific type of pain |
|---|---|---|---|---|---|---|
| [20] | 31 | | | Chronic | Musculoskeletal | Lower back |
| [21] | 319 | | 76.5% | | Musculoskeletal | Jaw |
| [22] | 15,715 | | | Acute | Musculoskeletal | Lower back |
| [23] | | | | | Musculoskeletal | Back |
| [24] | 121 | 45.0 | 60.0% | | Headache | Migraine, Cluster Headache |
| [25] | 65 | 56.4 | 61.5% | Chronic | Musculoskeletal | Osteoarthritis, Rheumatoid Arthritis, Spondylarthritis, Ankylosing Spondylitis |
| [26] | | | | Chronic | Musculoskeletal | Osteoarthritis |
| [27] | 92 | | | Chronic | Musculoskeletal | |
| [29] | 462 | | | | Cancer | Bone metastases |
| [30] | 40 | | | Acute | | Sickle Cell Disease |
| [31] | 87 | 24.2 | 72.4% | | Cancer (treatment) | |
| [32] | | | | | Musculoskeletal | Knee |
| [34] | 1,943 | | | | Musculoskeletal | Lower back |
| [35] | 42 | 45.3 | 40.3% | Chronic | Musculoskeletal | Back |
| [36] | 84 | 45.3 | 43.0% | Chronic | Musculoskeletal | Back |
| [37] | 576 | | | | Musculoskeletal | Knee, Hip |
| [38] | 135 | 41.1 | 54.0% | Chronic | Musculoskeletal | Back |
| [39] | 33 | 62.0 | | Chronic | Cancer | Prostate |
| [40] | 3,668 | 48.0 | 55.3% | Chronic | Noncancer | |
| [41] | 71 | 55.4 | 54.0% | Chronic | Musculoskeletal | Lower back, Osteoarthritis, Rheumatoid Arthritis, Fibromyalgia |
| [42] | 93 | 49.6 | 86.0% | Chronic | Nonterminal | |
| [44] | | | | Chronic | | |
| [45] | 5 | | | Chronic | Musculoskeletal | Lower back |
| [46] | 723 | | 47.0% | Acute, Chronic | | |
| [47] | 127,871 | 51.8 | 11.8% | Acute, Chronic | Musculoskeletal | |
| [48] | 11,416 | | 15.8% | | Musculoskeletal | |
| [49] | 364 | 53.7 | 62.7% | Chronic | Musculoskeletal | Lower back |
| [50] | 643 | 53.5 | 51.6% | | Musculoskeletal | Knee, Hip |
| [51] | | | | Acute, Chronic | Musculoskeletal | Fibromyalgia |
| [52] | 715 | | | Chronic | | |
| [53] | 709 | | | Chronic | | |
| [54] | | | 100.0% | Chronic | Noncancer | Endometriosis |
| [55] | 22 | | | Chronic | Musculoskeletal | Osteoarthritis, Rheumatoid Arthritis, Back, Psoriatic Arthritis, Fibromyalgia, Dish Syndrome |
| [58] | 123 | 14.7 | 78.9% | Chronic | | |
| [59] | | | | Acute, Chronic | | |

## Textual data

The studied languages were English (n=35), Finnish (n=1), Japanese (n=1), Dutch (n=1), Korean (n=1), and Portuguese (n=1). The studied language of pain was generated by physicians (n=20), patients (n=14), or both (n=6). Physician-generated data was obtained from electronic health records (EHRs; n=22) and interviews (n=3), whilst patient-generated data was obtained from interviews (n=11) and social media (n=7). The average reported number of documents used for analysis was 20,954 (Q2=626, Q3=3,426). Only a few studies reported the average length of their documents (n=6), with an average length of 500 tokens (Q2=374.9, Q3=888.5). Table A.3 in Supplementary Materials shows textual data availability.

Tab. 3 shows these data.

Table 3. Textual data.

| Ref. | Language | Type of pain language | Source | Number of documents | Average document length in tokens/words |
|---|---|---|---|---|---|
| [20] | English | Physicians | Electronic health records (EHRs) | 31 | |
| [21] | Korean | Physicians | EHRs | 319 | |
| [22] | English | Physicians | EHRs | 17,409 | |
| [23] | English | Patients | Social media | 104,274 | |
| [24] | Dutch | Patients | Interview | 121 | 476.0 |
| [25] | Portuguese | Patients | Interview | 65 | |
| [26] | English | Physicians | EHRs | | |
| [27] | English | Physicians | EHRs | 1,058 | |
| [28] | English | Physicians | EHRs | 702 | |
| [29] | English | Physicians | EHRs | 1,459 | 1038.3 |
| [30] | English | Physicians | EHRs | 424 | |
| [31] | English | Patients, Physicians | Interview | 391 | |
| [32] | English | Patients | Interview | 55 | |
| [34] | English | Physicians | EHRs | 2,749 | |
| [35] | English | Patients | Interview | 42 | |
| [36] | English | Patients | Interview | 84 | |
| [37] | English | Physicians | EHRs | 576 | |
| [38] | English | Patients | Interview | 135 | |
| [39] | English | Physicians | EHRs | 4,409 | |
| [40] | English | Physicians | EHRs | 3,668 | |
| [41] | English | Patients | Interview | 71 | 273.8 |
| [42] | English | Patients | Interview | 93 | |
| [43] | English | Physicians | EHRs | 3,750 | |
| [44] | English | Physicians | EHRs | 500 | |
| [45] | Finnish | Physicians | EHRs | 15 | |
| [46] | English | Physicians | EHRs | 1,985 | 1026.0 |
| [47] | English | Physicians | EHRs | 270,915 | |
| [48] | English | Physicians | EHRs | 63,812 | |
| [49] | English | Physicians | EHRs | 626 | |
| [50] | English | Physicians | EHRs | 586 | |
| [51] | English | Patients | Social media | 399 | |
| [52] | English | Patients | Social media | 941 | |
| [53] | English | Patients | Social media | 937 | 53.0 |
| [54] | English | Patients | Social media | 70,817 | |
| [55] | English | Patients, Physicians | Interview | 51 | |
| [56] | Japanese | Patients, Physicians | EHRs, Social media | 619 | |
| [57] | English | Patients, Physicians | Social media | 836 | |

| [58] | English | Patients, Physicians | Interview | 3,426 | 132.7 |
| [59] | English | Patients, Physicians | EHRs, Social media | 200 | |
| [60] | English | Patients | Social media | 216,873 | |

## Methodology

The computational methodology is structured similarly for all studies, including the preprocessing, main task processing, and evaluation stages. Table A.4 in Supplementary Materials shows the code availability. According to [17], the average scores per reporting quality item (1-9), were, respectively: 1, 0.7, 0.5, 0.9, 0.9, 0.3, 0.4, 0.5, 0.7. The two lowest scores were associated with reporting model bias and model parameters for reproducibility, which most studies lacked. The overall reporting quality score was 0.7.

## Preprocessing

All studies followed a similar preprocessing pipeline, including the removal of unwanted tokens/ entities (n=22), such as stop-words, infrequent tokens, punctuation, non-alphanumerical tokens, etc., and text normalization (n=21), such as anonymization, spelling errors, lemmatization, stemming, lower-casing, among others. These were all standard techniques for NLP methodologies. Table A.5. in Supplementary Materials shows these data.

## Main task

Biomedical entity extraction encompasses the identification and possible linkage of biomedical entities within a text to biomedical concepts in a pre-defined ontology. An unlinked entity is not immediately distinguishable from other unlinked entities of the same ontology class. Some studies defined their own task-specific ontologies/ dictionaries (n=3), whilst others used pre-defined, standard ontologies/dictionaries, such as UMLS (n=1), ICF (n=1), PCQ indicators (n=2) and SDoH (n=1). Some of these required the development of the corresponding ontology dictionary. For the extraction and possible linkage, some studies developed their own pipelines, such as rule-based, pattern matching (n=5), and deep learning-based (n=1), whilst others applied publicly or privately available pipelines, namely Clinical Language Annotation, Modeling, and Processing (CLAMP) [61] (n=1), Headai's Graphmind (HGM) [8] (n=1), clinical Text Analysis and Knowledge Extraction System (cTAKES) [62] (n=1), and MetaMap [63] (n=1). Table A.6 in Supplementary Materials shows these data.

The classification task was structured in a standard way, i.e., feature extraction followed by classification (or comparison of multiple classifiers). Extracted features were either domain-driven (n=6), such as biomedical entity extraction and linking based on cTAKES or MetaMap, and psycholinguistics based on Linguistic Inquiry and Word Count (LIWC) [64], or domain-agnostic (n=11), such as word frequency, bag-of-words (BOW), Term Frequency – Inverse Document Frequency (TF-IDF), word embeddings, and topic distribution. Classification was performed with both traditional models (n=12), such as rule-based, Logistic Regression (LR), Support Vector Machine (SVM), Decision Tree (DT), Random Forest (RF), and Naïve Bayes (NB), and deep learning models (n=5), such as Feed-Forward Neural Network (FFNN), Bidirectional Encoder Representations from Transformers (BERT)-based [65], and Convolutional Neural Network (CNN). Tab. A.7 shows these data. The universe of classes of each study is shown in Tab. A.8.

Correlation analysis was mainly structured in two stages: (1) feature extraction from the language of pain, and (2) feature correlation with target clinical parameters. Most studies used direct (psycho)linguistic features for their analysis, either with LIWC (n=2) or lower-level features, such as word frequency and topic importance (n=1). Other studies defined their features as inferred clinical states or risks (n=2).

---

[8] https://headai.com/

Correlation analysis mostly followed standard statistical approaches, based on univariate and multivariate regression analyses. Table A.9 in Supplementary Materials shows these data.

Lexico-semantic analysis, similarly to correlation, starts by extracting language-based features from the source language of pain, followed by analyzing distinctive characteristics. In general, there were two main approaches for feature extraction: (key)word frequencies and co-occurrence (n=7), and topic modelling, either manually (n=1), or with Latent Dirichlet Allocation (LDA) [66] (n=2), Differential Language Analysis (DLA) (n=1), or Leximancer [9] (n=1). Word-based features were analyzed with word clouds and network analyzes, whilst topic analysis was qualitative and comparative between sub-populations. Although focused on lexico-semantic analyses, two studies [59,60] had additional outcomes, namely a pain lexicon for downstream NLP tasks, and a proposal for an updated version of the MPQ, respectively. Table. A.10 in Supplementary Materials shows these data.

## Evaluation

Evaluation methods are intrinsically related to the main task. k-Fold cross validation (CV; n=13) was the main testing framework for supervised tasks. Classification and extraction resorted to commonly used supervised metrics, such as recall/ sensitivity (n=17), precision/ positive predictive value (PPV; n=15), $F_1$ (n=15), accuracy (n=8), area under the curve (AUC; n=6), specificity (n=4), and negative predictive value (NPV; n=1). Some of these works also resorted to qualitative evaluation methods (n=4). Correlation and lexico-semantic analysis resorted mostly to correlation and qualitative metrics, respectively. Table A.11 in Supplementary Materials shows these data.

## Outcome targets

Most studies targeted their outcomes at physicians (n=28) and only a few at patients (n=5). The dimensions of pain addressed by the studies were sensory (n=29), physiologic (n=24), cognitive (n=22), behavioral (n=20), affective (n=14), and sociocultural (n=9). The targeted stages of care were treatment/ rehabilitation (n=19), clinical-decision support (n=18), diagnosis (n=17), and self-management (n=6).

Table 4 shows these data. Table A.12 in Supplementary Materials shows each study's main outcome.

Table 4. Outcome targets. Dimensions of pain: P=Physiologic, S=Sensory, A=Affective, C=Cognitive, B=Behavioral, SC=Sociocultural

| Ref. | Main beneficiary | | Dimensions of pain | | | | | | Stage of care | | | |
|---|---|---|---|---|---|---|---|---|---|---|---|---|
| | Physician | Patient | P | S | A | C | B | SC | Diagnosis | Clinical-decision support | Treatment/ Rehabilitation | Self-management |
| [20] | ✓ | | ✓ | | | | | | ✓ | | | |
| [21] | ✓ | | ✓ | ✓ | | ✓ | ✓ | | ✓ | | | |
| [22] | ✓ | | ✓ | ✓ | | | | | ✓ | ✓ | | |
| [23] | | ✓ | | ✓ | | ✓ | | | | | | ✓ |
| [24] | ✓ | | ✓ | ✓ | ✓ | | | | ✓ | | | |
| [25] | | | | ✓ | | | | | ✓ | | | |
| [26] | ✓ | | | ✓ | | | | | ✓ | ✓ | | |
| [27] | | | ✓ | ✓ | ✓ | | | | ✓ | ✓ | | |
| [28] | | | | ✓ | | | | | | | ✓ | |
| [29] | ✓ | | | ✓ | | | | | ✓ | ✓ | | |
| [30] | ✓ | | | ✓ | | | | | ✓ | | | |
| [31] | ✓ | | ✓ | ✓ | | ✓ | | ✓ | | ✓ | ✓ | |

---

[9] https://www.leximancer.com/

| | | | | | | | | | | | | |
|---|---|---|---|---|---|---|---|---|---|---|---|---|
| [32] | ✓ | | ✓ | ✓ | ✓ | ✓ | ✓ | | | ✓ | ✓ | |
| [34] | ✓ | | ✓ | | | ✓ | | | ✓ | ✓ | | |
| [35] | ✓ | | ✓ | ✓ | ✓ | | | | | | ✓ | |
| [36] | ✓ | | | | | ✓ | | | | | ✓ | |
| [37] | ✓ | | ✓ | ✓ | | ✓ | ✓ | | | ✓ | ✓ | |
| [38] | | | | ✓ | ✓ | ✓ | | | | | ✓ | |
| [39] | ✓ | | ✓ | ✓ | | | | | ✓ | ✓ | | |
| [40] | ✓ | | | | ✓ | ✓ | | | | | ✓ | |
| [41] | ✓ | | | | ✓ | ✓ | | | | | ✓ | |
| [42] | | | | ✓ | ✓ | ✓ | ✓ | | | ✓ | | |
| [43] | ✓ | | ✓ | ✓ | ✓ | | | ✓ | | ✓ | | |
| [44] | | | ✓ | ✓ | ✓ | ✓ | ✓ | ✓ | ✓ | ✓ | | |
| [45] | ✓ | | ✓ | | | ✓ | ✓ | | | ✓ | ✓ | |
| [46] | | | ✓ | ✓ | | ✓ | | | ✓ | ✓ | | |
| [47] | ✓ | | ✓ | ✓ | ✓ | ✓ | | | | | ✓ | |
| [48] | ✓ | | ✓ | ✓ | ✓ | ✓ | ✓ | | ✓ | ✓ | ✓ | |
| [49] | | | | ✓ | ✓ | | | | ✓ | | | |
| [50] | ✓ | | ✓ | | | | | | | ✓ | | |
| [51] | ✓ | | ✓ | ✓ | ✓ | ✓ | ✓ | ✓ | | ✓ | ✓ | |
| [52] | | | ✓ | ✓ | ✓ | ✓ | | | | | | |
| [53] | ✓ | | ✓ | ✓ | | ✓ | ✓ | | | | ✓ | ✓ |
| [54] | ✓ | ✓ | ✓ | ✓ | ✓ | ✓ | ✓ | ✓ | | | ✓ | ✓ |
| [55] | ✓ | ✓ | | ✓ | ✓ | ✓ | ✓ | ✓ | | | ✓ | ✓ |
| [56] | ✓ | ✓ | ✓ | | | ✓ | ✓ | | | | ✓ | ✓ |
| [57] | ✓ | | ✓ | | ✓ | | ✓ | | | ✓ | ✓ | |
| [58] | ✓ | ✓ | | | ✓ | ✓ | ✓ | ✓ | | | ✓ | ✓ |
| [60] | | | | ✓ | ✓ | ✓ | | | ✓ | | | |

# DISCUSSION

The review research questions explore pain language analysis via computational methods, covering its prevalence, tasks linked to patient- and physician-generated language of pain, integration of linguistic findings, primary beneficiaries, pain dimensions, targeted pain care stages, and the impact of proposed algorithms on physician performance in pain-related clinical tasks. The following discussion addresses each of these questions.

## Which language type is most analyzed? (Q1.a)

Physician-generated language of pain, specifically from EHRs, was the most used type of language. It is usually large and accompanied by structured patient metadata, which can provide both contextualization for controlling variables and expert annotated labels, such as International Classification of Diseases (ICD) coding. The free text itself is generated in a clinical domain, with the specific aim of describing clinical parameters [67]. These data are also more commonly publicly available in standard datasets than patient-generated data [68,69].

Patient-generated language has a more colloquial lexical profile and may lack some of the physician-generated data characteristics, depending on the collection conditions. Patient interviews allow for a granular control over the data that is captured, namely controlling variables, expert annotated labels, and contextual aim. However, such collection protocols are difficult to implement and may have limited reach, both in size and applicability. Social media documents, on the other hand, have a large public availability, but there is little control over the data that is collected and the contextual aim of the document, usually requiring some assumptions (e.g., self-reported diagnoses) [70].

Importantly, although both physician- and patient-generated languages describe the same object (i.e., the experience of pain), only the patient has direct access to the object itself (i.e., the personal experience).

## Which tasks are performed for each language type? (Q1.b)

For classification-based studies, questionnaire automation and treatment response were performed only with patient-generated language of pain [32,35,36]. These tasks, as presently formulated, could not be performed with any other source of language of pain. Similarly, pain assessment and risk factor identification were performed only with physician-generated language of pain, because no other source would be adequate [27,34]. Patient diagnosis/ phenotyping and PRO identification used both sources of language of pain for similar outcomes [20–26,31]. Finally, pain mention identification and patient triaging only used physician-generated data [28–30,37]. However, it seems reasonable to also perform these based on patient-generated data, and thus, we conclude that there is a gap in this line of research. One possible challenge, especially for patient triaging based on patient-generated language, is in assuming that patient concerns are relevant for that task, especially when the patients themselves are unaware of their clinical status (otherwise, triaging would not be necessary). Arguably, from the studies performing lexico-semantic analysis of patient-generated language of pain, patient concerns are clinically relevant [51]. However, in such studies, the included patients assumingly already knew their clinical status, and, therefore, were already biased towards specific concerns. Thus, it is not clear from the reviewed literature if patient concerns are clinically relevant for triaging when they are unaware of their clinical status.

Entity extraction was only performed with physician-generated language of pain, leaving this line of research clearly lacking in entity extraction from patient-generated language. One possible challenge for the extraction (and linkage) of biomedical entities from patient-generated language might be the adaptation of existing tools, designed for biomedical or clinical text (e.g., MetaMap, cTAKES, CLAMP), to colloquial language, or even the development of new ones. Indeed, no study directly applied these tools to patient-generated language of pain, so the baseline performance for this adaptation is also lacking.

Lexico-semantic analysis was addressed with patient-generated language of pain, although some studies also included physician-generated data [55–58]. Their purpose was to compare linguistic characteristics between patients and physicians when describing pain management, which is an inherently collaborative task. Similarly, correlation analysis mostly observed correlation between linguistic patient features and clinical parameters, specifically because the research aims were to identify linguistic clues for the patient clinical status.

## Are previous linguistic findings explicitly incorporated into the main processing? (Q2)

Only one study [60] explicitly incorporated the linguistic findings of [5] (MPQ). The grammatical findings by [6] and [7] were never explicitly incorporated, nor referenced. The reason for the widespread exclusion of these linguistic findings is unclear. Importantly, these are specifically applied to patient-generated language of pain, and most studies in this review do not use such type of language. Another possible explanation is that their inclusion may require the use of algorithms taking advantage of rules and/or grammars, which can be laborious to define. However, since their clinical correlation is lacking, it is currently unknown if their exchange for easier-to-implement deep learning approaches is reasonable. Moreover, the inclusion of these linguistic findings to address the various tasks mentioned in this review has potentially positive outcomes. First, they already reveal phrasal structures that indicate certain aspects of the pain experience, supported by linguistic theories and ample data. Second, the data may already include contents equivalent to these phrasal structures, becoming easier to leverage. Finally, because they are focused on patient-generated language, they could help with the adaptation or development of clinical text tools for colloquial language, facilitating the challenges previously identified.

## Which dimensions of pain are most and less studied? (Q4)

The most studied pain dimensions are physiologic, sensory, cognitive, and behavioral. These were studied either directly from the language of pain, or indirectly as controlling variables or post-hoc analyzes. When directly studied from the language of pain, physiologic and sensory aspects of pain, e.g., location, duration, and intensity, were commonly extracted from the text, or used as prediction target. Cognitive and behavioral aspects, on the other hand, were commonly integrated in thematic analyzes of the language of pain.

The least studied are the affective and sociocultural dimensions. Indeed, not many studies took advantage of off-the-shelf sentiment analysis tools to assess patient mood and well-being, although text can be a good indicator of sentiment [71]. The sociocultural dimension, on the other hand, is harder to gauge, but it can also manifest linguistically, namely in word-choice, sentence construction, and focused themes. Moreover, the pain experience can be modulated by the sociocultural context in which it is described, e.g., a clinical context may invite the use of certain terms that the patient would otherwise not use, and vice-versa. Indeed, no study compared the linguistic characteristics and clinical relevance of patient-generated language of pain in a clinical context versus in-the-wild (i.e., colloquial language). This is especially relevant for the design of patient-centered tools aimed to be used in daily contexts.

## For whom are the studies, and their outcomes primarily designed, and which stage of pain care is being targeted? (Q3, Q5)

Most of the studies included in this review were targeted for physicians, either directly as useful clinical insights, or as tools that can eventually be integrated in the clinical care pipeline. Very few studies were targeted at patients [23,54–56,58], which is closely related with the targeted stage of care: it seems that only a few outcomes were patient-focused because patients are mostly involved on the self-management stage, which is by far the least targeted one. Treatment/rehabilitation is the most targeted stage of care. Due to the recurrent care necessary for pain treatment, especially for chronic pain, this observation seems to be in line with clinical needs. The same applies for clinical-decision support, which is the second most targeted stage of care. These, together with the diagnosis stage, also seem to be the easiest to perform in terms of data availability, namely EHRs with clinical annotations and longitudinal assessments. On the other hand, self-management requires the active inclusion of patients in the study, which limits data abundance and availability. We believe this is one of the reasons why this is the least targeted stage of care.

We argue that there is potential in targeting patients and pain self-management, specifically with computational processing and NLP. As stated elsewhere, after sufficient time, pain can become a daily partner, with variations dependent on internal and external aspects. A tool designed to be always available to patients, providing context for their experience considering these aspects (e.g., how they feel today in comparison to yesterday or the previous week), has the potential to greatly improve patient empowerment. This may increase control over pain, which, over time, helps patients cope with their clinical status and make the best out of their situation. Therefore, we conclude that there is a research gap in this area and more work is necessary with patients as the primary beneficiaries.

## Does physician performance on clinical tasks improve with the inclusion of the proposed algorithm? (Q6)

Physicians learn to communicate with patients and extract clinically relevant information from that exchange. Years of experience allow them to develop mental models for all dimensions of pain and the way they manifest in in-person consultations [72]. If the tools developed in this line of research (i.e., computational pain assessment from patient narratives) are to be included in the pain care pipeline, their performance should be compared to that of physicians for the same task, under identical conditions. Moreover, since the aim of these tools is to assist physicians (as opposed to replacing them), physician performance gains when assisted by these tools should also be measured. Importantly, various studies partnered with experts and physicians to create the golden annotation on which they then measured the performance of their tools. This alone, however, does not allow measuring the tool performance in comparison to physician performance. For a given task, this requires the golden annotation, the tool itself, and an ensemble of physicians unfamiliar with the test set. Both tool and physician performance should be measured based on the same test set input and golden annotation. Only one study included in this review compared these performances and their gains, finding that the best performance for their task was obtained by combining the efforts of both algorithm and clinicians [45]. Thus, there is still a lack of evidence for the practical advantage of using these tools.

## Limitations

This systematic review is highly focused on the merger of two domains: NLP and pain. Although we tried to include every relevant keyword from each domain, to accommodate for the maximum number of studies in this research field, this type of database search is still limited by definition. By excluding articles based on their written language, we may also be overlooking relevant developments. Additionally, although we based this review on other closely related reviews, it is important to note that there are no other directly comparable data extraction and synthesis methodologies available, which represents an important limitation.

# CONCLUSIONS

This systematic review delves into the intersection of NLP and the study of pain narratives, examining research works that computationally analyze the language of pain generated by patients or physicians. It is structured around a set of well-defined review research questions. These questions cover important aspects, such as the type of language of pain analyzed, tasks performed, incorporation of linguistic findings, target audience, dimensions of pain studied, and addressed stages of clinical pain care. We found a limited exploration of patient-generated language of pain and incorporation of linguistic findings, limited development of self-management and patient-centered tools, limited exploration of affective and sociocultural dimensions, and practically inexistent measurement of gains in physician performance on clinical tasks when aided by the proposed tools. Importantly, to our knowledge, there is no previously published systematic review that defines the landscape of research at the intersection of NLP and pain.

# AUTHOR CONTRIBUTIONS

Search keyword definition, article selection, and data extraction were performed by all authors. Diogo A.P. Nunes performed article retrieval, developed data synthesis, and wrote this article, all of which were then reviewed and validated by all authors.

# FUNDING

Diogo A.P. Nunes is supported by a scholarship granted by Fundação para a Ciência e Tecnologia, with reference 2021.06759.BD. This work was supported by Portuguese national funds through Fundação para a Ciência e Tecnologia, with reference UIDB/50021/2020. This research was supported by the Portuguese Recovery and Resilience Plan (RRP) through project C645008882-00000055 (Responsible.AI).

## CONFLICTS OF INTEREST

The authors of this systematic review are the authors of one of the included studies [25]. There are no other conflicts of interest to declare.

## ACRONYMS

| | |
|---:|---|
| AUC | Area Under the Curve |
| BOB | Bag-of-Bigrams |
| BOW | Bag-of-Words |
| CNN | Convolutional Neural Network |
| CUI | Concept Unique Identifier |
| CV | Cross Validation |
| DLA | Differential Language Analysis |
| DT | Decision Tree |
| EHR | Electronic Health Record |
| FFNN | Feed Forward Neural Network |
| ICD | International Classification of Diseases |
| ICF | International Classification of Functioning, Disability, and Health |
| kNN | k-Nearest Neighbors |
| KOOS | Knee injury and Osteoarthritis Outcome Score |
| LBP | Low back pain |
| LDA | Latent Dirichlet Allocation |
| LR | Logistic Regression |
| LSA | Latent Semantic Analysis |
| MeSH | Medical Subject Headings |
| MPQ | McGill Pain Questionnaire |
| NB | Naïve Bayes |
| NER | Named Entity Recognition |
| NLP | Natural Language Processing |
| NPV | Negative Predictive Value |
| PCQ | Pain Care Quality |
| POS | Part-of-Speech |
| PPV | Predictive Positive Value |
| PRC | Precision-Recall Curve |
| PRO | Patient-Reported Outcome |
| PROM | Patient-Reported Outcome Measurement |
| RegEx | Regular Expression |
| RF | Random Forest |
| ROC | Receiver Operating Characteristic |
| SDoH | Social Determinants of Health |
| SNOMED-CT | Systematized Nomenclature of Medicine – Clinical Terms |
| SVM | Support Vector Machine |
| TF-IDF | Term Frequency – Inverse Document Frequency |
| TMD | Temporomandibular |
| UMLS | Unified Medical Language System |
| VDP | Verbally Declared Pain |
| VHA | Veteran's Health Administration |
| W2V | Word2Vec |

## FIGURE LEGENDS

Figure 1: PRISMA chart for study selection.

Figure 2: Distribution of selected studies per publication year.

# SUPPLEMENTARY MATERIAL

# Computational analysis of the language of pain: a systematic review

**Diogo Afonso Pedro Nunes (Diogo A.P. Nunes) – corresponding author**
[1] Instituto de Engenharia de Sistemas e Computadores - Investigação e Desenvolvimento, Rua Alves Redol, 9, 1000-029 Lisboa, Portugal.
[2] Instituto Superior Técnico, Universidade de Lisboa, Avenida Rovisco Pais, 1049-001 Lisboa, Portugal.
diogo.p.nunes@inesc-id.pt
0000-0002-6614-8556

**Joana Maria de Pinho Ferreira Gomes (Joana Ferreira-Gomes)**
[3] Instituto de Investigação e Inovação em Saúde da Universidade do Porto (I3S). Rua Alfredo Allen 208, 4200-393 Porto, Portugal.
[4] Instituto de Biologia Molecular e Celular (IBMC), Universidade do Porto. Rua Alfredo Allen 208, 4200-393 Porto, Portugal.
[5] Departamento de Biomedicina – Unidade de Biologia Experimental, Faculdade de Medicina, Universidade do Porto. Alameda Prof. Hernâni Monteiro 4200-319 Porto, Portugal.
jogomes@med.up.pt
0000-0001-8498-6927

**Fani Lourença Moreira Neto (Fani Neto)**
[3] Instituto de Investigação e Inovação em Saúde da Universidade do Porto (I3S). Rua Alfredo Allen 208, 4200-393 Porto, Portugal.
[4] Instituto de Biologia Molecular e Celular (IBMC), Universidade do Porto. Rua Alfredo Allen 208, 4200-393 Porto, Portugal.
[5] Departamento de Biomedicina – Unidade de Biologia Experimental, Faculdade de Medicina, Universidade do Porto. Alameda Prof. Hernâni Monteiro 4200-319 Porto, Portugal.
fanineto@med.up.pt
0000-0002-2352-3336

**David Manuel Martins de Matos (David Martins de Matos)**
[1] Instituto de Engenharia de Sistemas e Computadores - Investigação e Desenvolvimento, Rua Alves Redol, 9, 1000-029 Lisboa, Portugal.
[2] Instituto Superior Técnico, Universidade de Lisboa, Avenida Rovisco Pais, 1049-001 Lisboa, Portugal.
david.matos@inesc-id.pt
0000-0001-8631-2870

**Tables**: 12
**Figures**: 0
List of acronym definitions at the end.

Table A.1. Search keywords, for article retrieval, for each of the axes in the systematic review.

| Axis | Keywords |
|---|---|
| Computational processing | machine learning, deep learning, natural language processing, nlp, text mining, text processing, text analysis |
| Language of pain | language, linguistics, clinical text, clinical narrative, clinical note, patient text, patient narrative, patient note, biomedical text, biomedical narrative, biomedical note, medical text, medical narrative, medical note, pain, chronic pain, painful |

Table A.2. Extracted data points.

| Group | Data points |
|---|---|
| Overview | Article title<br>Publishing year<br>Primary purpose<br>Primary task |
| Patient population | Number<br>Average age<br>Percentage of female-gendered<br>Chronicity of pain<br>Generic type of pain<br>Specific type of pain |
| Textual data | Language<br>Type of pain language (patient- or physician-generated)<br>Source (HER, interview, social media)<br>Number of documents<br>Average document length (in tokens)<br>Data availability |
| Methodology | Textual data preprocessing<br>Main task processing<br>Evaluation method(s)<br>Code availability |
| Outcome targets | Primary outcome question<br>Primary outcome answer<br>Target(s) (physicians, patients)<br>Dimension(s) of pain (physiologic, sensory, affective, cognitive, behavioral, or sociocultural) [1]<br>Stage(s) of care (diagnosis, clinical-decision support, treatment/rehabilitation, or self-management) |

Table A.3. Textual data availability.

| Ref. | Availability | Availability access |
|------|--------------|---------------------|
| [2] | Available on request | |
| [3] | Available on request | |
| [4] | Available on request | |
| [5] | Available | i2b2, available: https://portal.dbmi.hms.harvard.edu/projects/n2c2-nlp/; MIMIC-III, available: https://physionet.org/content/mimiciii-demo/1.4/; ARIA: not stated |
| [6] | Available on request | |
| [7] | Not available | |
| [8] | Available | https://github.com/yonestar/Ashar_2023_CBP_reattribution |
| [9] | | Vocabulary and rules are available on supplementary data |
| [10] | Available on request | |
| [11] | Available on request | |
| [12] | Not available | |
| [13] | Available on request | |
| [14] | Available on request | |
| [15] | Available on request | |
| [16] | Not available | |
| [17] | Partially available | CRIS, available upon request: https://projects.slam.nhs.uk/research/cris#:~:text=It%20provides%20authorised%20researchers%20with,SLaM)%20electronic%20clinical%20records%20system; MIMIC-III, available: https://physionet.org/content/mimiciii-demo/1.4/; Reddit & Twitter: Not stated. |

Table A.4. Code availability.

| Ref. | Availability | Availability access |
|---|---|---|
| [18] | Available | https://github.com/haven-jeon/KoNLP |
| [5] | Available | https://github.com/hn617/texTRACTOR |
| [6] | Available | https://github.com/zhaohualu/nlp4pro1 |
| [7] | Available on request | |
| [9] | Not available. Suggested open-source modules. | |
| [19] | Available on request | |
| [10] | Available | https://github.com/jayachaturvedi/pain_in_mental_health |
| [12] | Available | https://github.com/BCHSI/social-determinants-of-health-clbp |
| [20] | Available | https://www.leximancer.com/ |
| [17] | Available | https://github.com/jayachaturvedi/pain_lexicon |

Table A.5. Preprocessing techniques.

| Ref. | Remove | Normalize | Other |
|---|---|---|---|
| [21] | | ✓ anonymization | |
| [18] | ✓ tokens with count < 2 | ✓ manually - correction of spelling errors, categorization of symptoms | ✓ manual categorization of synonyms |
| [2] | ✓ punctuation, numbers, links, emails, dates | ✓ tokens with count < 5 with token with minimum edit distance, lemmatization | |
| [3] | ✓ duplicates, repeated characters, non-English tweets | ✓ links to images with special token, links to retweets with special token, other links with special token, emoticons with corresponding text | |
| [4] | ✓ stop-words | ✓ manual cleaning | |
| [22] | | ✓ lemmatization | ✓ POS tagging, punctuation inference, transcription |
| [23] | ✓ duplicates, special characters, punctuation | ✓ POS tagging, NER tagging | |
| [24] | ✓ stop-words | ✓ stemming | |
| [25] | ✓ duplicates | ✓ lower-case | |
| [5] | ✓ whitespaces, special characters | ✓ medical abbreviations with expansion, lower-case | |
| [26] | ✓ stop-words, punctuation | ✓ lemmatization | |
| [7] | | | ✓ POS tagging, syntactic dependencies |
| [27] | | | |
| [28] | ✓ stop-words | ✓ lemmatization, tokens with embedding index | ✓ POS tagging, LIWC [29] tagging |
| [30] | ✓ stop-words | ✓ lemmatization, tokens with embedding index | ✓ POS tagging, LIWC tagging |
| [31] | ✓ punctuation, special characters | ✓ personal names with special token, subset of phrases with special token, contractions with expansion, abbreviations with expansion, lower-case | |
| [8] | ✓ stop-words, typographical errors, tokens with length < 2 or length > 20 | ✓ lemmatization | |
| [9] | ✓ identifiers | ✓ dates | |
| [32] | | ✓ reports of pain with x/10 | |
| [19] | ✓ punctuation, determiners, title abbreviations | ✓ lower-case | |
| [10] | | ✓ anonymization | |
| [33] | ✓ punctuation, irrelevant characters | ✓ lower-case | |
| [12] | ✓ stop-words | ✓ lemmatization | |
| [13] | | ✓ entities with UMLS [34] semantic types | |
| [35] | ✓ POS != noun | ✓ POS tagging | |
| [15] | ✓ stop-words, links, punctuation, internet symbols | ✓ lemmatization, lower-case | |
| [16] | ✓ stop-words, words not in the same document as word "pain" | ✓ POS tagging | |
| [36] | ✓ stop-words | | |
| [37] | ✓ manually - usernames, links, internet symbols, irrelevant contexts, automatically - posts with CrystalFeel [38] sarcasm score > 0.7 | | |

Table A.6. Main extraction methodology.

| Ref. | Ontology definition (and sample annotation) | Entity extraction (and linkage) |
|---|---|---|
| [32] | 1. Create pain ontology with focus group, with reference to CUIs.<br>2. Sample annotation for training and validation. | 1. Apply CLAMP [39]. |
| [19] | 1. Human annotation of training sample using pre-defined ontology (mapping exemplars, i.e., unique text strings, to ontology classes).<br>2. Create ontology dictionary: map each training exemplar to a corresponding BOW, BOB, and frequency count within that exemplar's class (it may appear in other classes). | 1. Convert each test document to (1-5)-grams, each of which is also converted into BOW and BOB (identical to the exemplar dictionary).<br>2. Compare the BOW and BOB of each n-gram of each test document with every single entry in the exemplar dictionary of size n to 4n, using the Jaccard Index.<br>3. Select the top-n Jaccard Index exemplars for each test document n-gram.<br>4. Define output vector of each word in the test document as the sum of the scores of that word for each class of the exemplar dictionary (score is a combination of the Jaccard index and the frequency counts of the matching exemplar). Top-scoring class is the selected entity. |
| [40] | 1. Map each relevant ICF code (title and description levels) with a list of terms and phrases found on the data. | 1. Apply HGM [1] with 12 setups against medical expert annotation (each setup is a combination of the method (normal or fuzzy), the ICF coding level (title or description), and the augmentation of the initial mapping with more terms, from MeSH and/or "real life" (as determined by medical experts). |
| [10] | 1. Define pain ontology.<br>2. Use Pain Lexicon [17] as dictionary for pain mentions. | 1. Find mentions of pain using the Pain Lexicon dictionary.<br>2. Manually annotate pain mentions for pain relevance, anatomy, character, and management. (Although manual extraction, it was included due to its use of Pain Lexicon and explicit design for NLP integration. It is equated to a rule-based pipeline.) |
| [33] | 1. Sample annotation for PCQ indicators, information about the treatment of pain, and contextual modifiers.<br>2. Curate a normalized vocabulary using the terms found during annotation and from other standardized sources (UMLS, VHA). | 1. Identify targeted PCQ indicators in each sentence of each document using the curated vocabulary (not described). |
| [11] | 1. Same as [33]. | 1. Apply existing NLP algorithm to extract PCQ indicators.<br>2. Compare patterns of indicators across different visit types.<br>3. Evaluated co-occurrence of indicators. |
| [12] | 1. Create SDoH ontology, with additional anxiety, depression, and pain score classes.<br>2. Training sample annotation. | 1. Compare performance of 3 extraction approaches: cTAKES [41], deep learning, and hybrid LR with pattern matching. |
| [13] | 1. Use UMLS as ontology, restricted to a specific set of semantic types. | 1. Extract demographic and clinical data from referral letter, such as: age, gender, postcode, body mass index, medication, smoking status, and comorbidities (not described).<br>2. Extract and link biomedical entities to UMLS concepts from each referral letter using MetaMap [42].<br>3. Count mentions of each extracted concept and semantic type. |

---

[1] https://headai.com/

Table A.7. Main classification methodology.

| Ref. | Feature extraction | Classification |
|---|---|---|
| [21] | 1. Biomedical entity extraction and CUI linkage with negation, using a custom cTAKES pipeline.<br>2. Vectorization of each document according to normalized CUI frequency (and negation), in 2 ways: set of all possible CUIs, or a manually curated set of CUIs directly related to back pain. | 1. Bernoulli NB, Multinomial NB, LinearSVC, and Perceptron. |
| [18] | 1. Statistical analysis to compare word frequency between conditions.<br>2. Vectorize each patient according to "diagnostic clues" from linguistic and clinical analysis (not described). | 1. Multivariate LR. |
| [2] | 1. Model 1: TFI-DF (1-5-grams).<br>2. Model 2: TFI-DF (1-5-grams) with select words.<br>3. Model 3: Pre-train W2V word embeddings.<br>4. Model 4: Define set of keywords.<br>5. Model 5: Topic modelling with manual review and selection of the most appropriate topics.<br>6. Model 6: N/A. | 1. Model 1: LR.<br>2. Model 2: LR.<br>3. Model 3: CNN.<br>4. Model 4: Keyword search.<br>5. Model 5: Topic with greatest importance.<br>6. Model 6: Notes with ICD-10 annotation defined as positive acute LBP. |
| [3] | N/A | 1. Fine-tune RoBERTa [43] with a classification layer. |
| [4] | 1. Select phrases manually labelled as "attack descriptions".<br>2. BOW ((1-3)-gram) vectorization (on the word and character level).<br>3. Patient metadata features (age and gender). | 1. Binary classification (SVM, LR, NB) of combinatorial feature vectors (BOW and metadata) for a total of 9 models. |
| [22] | 1. TF-IDF with and without POS tagging.<br>2. Topic distribution.<br>3. Early Fusion.<br>4. Late Fusion. | 1. DT, SVM |
| [23] | 1. Develop dictionary for 3 pain intensity classes (mild, moderate, severe).<br>2. Extract GloVe [44] embeddings for all dictionary and document words. | 1. Assign pain intensity class according to document and dictionary embedding cosine similarity. |
| [24] | 1. BOW (1-gram) vectorization | 1. kNN, DT, SVM, RF |
| [25] | 1. Given a positive document, using the Smith-Waterman alignment algorithm, find a token alignment with every other positive document, obtaining a list of phrases and their corresponding similarity score.<br>2. Given the list of phrases, generate a list of keys (not described).<br>3. For each key, produce 2 RegEx (one with distance control and other without).<br>4. Filter out generated RegEx based on a precision threshold (default is 100%).<br>5. Use resulting RegEx to match expressions in the documents, each match is given a prediction score which favors lengthier matches. | 1. Classifier chooses label according to highest scoring match.<br>2. If there is no match resort to other two classifiers: ALIGN and SVM. ALIGN simply selects the label of the text snippet with the highest similarity score given by the Smith-Waterman alignment algorithm. SVM is trained on BOW. |
| [5] | 1. Define look-up table with UMLS codes from the "signs and symptoms" semantic type relating to pain.<br>2. Define look-up table with exceptional, hypothetical, conditional, and historical keywords that might affect the semantics of the medical concept, expanding them with 5 more words per keyword using GloVe.<br>3. Extract and link biomedical entities to ICD codes from each phrase in a given sentence (in this case, sentence is a document), with the corresponding confidence score and negation, using MetaMap.<br>4. Identify all phrases with ICD codes relating to "pharmacologic substance".<br>5. Identify all ICD codes relating to the "signs and symptoms" UMLS semantic type – if a single phrase has more than one, select the one with the largest confidence score. | 1. Classify pain score based on rules and look-up tables.<br>2. Define VDP classification of a given note (multiple sentences) as the weighted average pain score of its sentences. |
| [26] | 1. Count (1-2)-gram frequency overall and per class (for both binary and multi-class settings).<br>2. Topic modelling with LDA [45] with k=2 (coherence score to determine k) | 1. Binary and multi-class classification with n-gram frequency and topic distributions as features (combinatorial nature for multiple setups), using various classifiers: LR, DT, RF, FFNN. |
| [6] | N/A | 1. End-to-end classification with BERT [46] (using a pre-trained model, further fine-tuned with their data). |
| | | 2. XGBoost and SVM classification using W2V word-embeddings (using a pre-trained model, further fined-tuned with their data).<br>3. Use as baseline TF-IDF+SVM, GloVe+SVM, GloVe+XGBoost, BioBERT [47], Blue-BERT, and ClinicalBERT [48]. |
| [7] | 1. Biomedical entity extraction and linking with MetaMap (medical conditions, surgical procedures, and anatomical structures), with negation detection.<br>2. Manually curated list of pain modifiers sorted for severity.<br>3. Sentiment polarity algorithm using StanfordCoreNLP [49] library. | 1. For Q1 and Q2 (multiple choice questions), use feature vector as is.<br>2. For Q3-Q10 (Likert-scale questions), use WEKA [50] for classification of ordinal classes (not described). |
| [27] | 1. Down-sampling of the majority class. | 1. Propose novel architecture (BERT-CNN) to handle large contexts. |
| [28] | 1. Extract 348 features: linguistic, psycholinguistic (LIWC), sentiment (VADER [51]), and semantic proximity (vector similarity based on LSA [52]). | 1. SVM and LR with L1 and L2 regularization and feature selection (grid search).<br>2. Linear regression for feature importance analysis. |
| [30] | 1. Integrate findings of previous study [28].<br>2. Extract 3 LIWC features: "Drives", "Achievement", and "Leisure".<br>3. Compute median value of each word proximity to 8 "topics": "magnify", "afraid", "fear", "awareness", "loss", "identity", "stigma", and "force". Proximity is calculated with the dot product between the vector of the word and the vector of the "topic". Vectors (300 dimensions) are obtained by applying LSA on a very large dataset. | 1. Fit binary classification model (linear regression) to study1 data (development set), applying bidirectional stepwise selection to further reduce number of features.<br>2. Apply resulting model to study2 data (test only). |
| [31] | 1. Use MetaMap to extract and link biomedical entities and replace them with their corresponding UMLS semantic type.<br>2. Topic modelling with LDA (1-gram, because preprocessing and previous step already aggregated relevant n-grams), with k=11, according to perplexity and four topic coherence metrics. | 1. Binary machine learning model for each class (not described). |

Table A.8. Universe of classes for classification-based studies.

| Ref. | Number of classes | Class names |
|---|---|---|
| [21] | 4 | "no back pain", "compressed nerve", "disc pain", "non-mechanical" |
| [18] | 2 | "TMD-mimicking", "actual TMD" |
| [2] | 2 | "acute low back pain", "other" |
| [3] | 2 | "complaint of back pain", "other" |
| [4] | 2 | "migraine", "cluster headache" |
| [22] | 3 | "mild", "moderate", "severe" |
| [23] | 3 | "mild", "moderate", "severe" |
| [24] | 2 | "has pain assessment mention", "does not have" |
| [25] | 3 | "have pain", "no pain", "other" |
| [5] | 3 + 3 | "irrelevant", "0 pain", "1 pain" + "pain", "no pain", "no mention of pain" |
| [26] | 2 + 4 | "pain relevant", "not" + "pain increase", "pain uncertain", "pain unchanged", "pain decrease" |
| [6] | 4 | "physical", "cognitive", "social", "unclassified" |
| [7] | 3 | 1, 2, 3 |
| [27] | 7 | "cancer", "weight", "fever", "infection", "bowel", "abnormal reflexes", "no risk factor" |
| [28] | 2 | "responder", "non-responder" |
| [30] | 2 | "responder", "non-responder" |
| [31] | 9 | "orthopedic referral", "discharge", "injection", "nutritionist", "physiotherapy", "diagnostic imaging surgery", "review appointment", "other" |

Table A.9. Main correlation analysis methodology.

| Ref. | Feature extraction | Analysis |
|---|---|---|
| [8] | N/A | 1. Identify and measure words with greatest frequency change from pre- to post-treatment.<br>2. Topical analysis using a text-scaling algorithm.<br>3. Dictionary-based (medical expert derived) automated attribution scoring algorithm. |
| [9] | 1. Define vocabulary and rules for pattern-matching and rating of each document (pain severity and clinical concepts).<br>2. Find mentions of UMLS body location concepts with ClinREAD [53], and replace with preferred term.<br>3. Find mentions of 16 UMLS semantic types with ClinREAD and replace with preferred term.<br>4. Contextualize mentions with negations, conditionals, and dates.<br>5. Cleanup processing into structured data of CUIs. | 1. Find pain severity (as extracted from the NLP pipeline) correlates from structured demographic and clinical data, in a univariate analysis.<br>2. Multiple regression model to assess the strength of associations between the occurrence of severe pain and all defined variables for which the p-value was significative in the univariate analysis.<br>3. Color-coding of longitudinal pain severity development per patient. |
| [54] | 1. Develop sets of keywords and phrases for each outcome and domain of the Opioid Risk Tool.<br>2. Augment sets with specific terms using cTAKES.<br>3. Apply RegEx-based algorithm (based on the defined sets of keywords and phrases) to EHRs (structured and unstructured parts) to score all elements of the Opioid Risk Tool.<br>4. For each patient, select the maximum score obtained in each Opioid Risk Tool element (multiple EHRs per patient).<br>5. Aggregate Opioid Risk Tool scores and convert to one of 3 classes – 0-3: low risk; 4-7: moderate risk; ≥ 8: high risk.<br>6. Binary classification to determine if patient was going to violate opioid agreement (not described). | 1. Correlate predicted risk for opioid agreement violation with actual agreement outcome. |
| [55] | 1. Extract 8 LIWC features: first person singular pronouns, pronouns referencing other people, anxiety, anger, sadness, positive emotions, causation, and insight. | 1. Correlation analysis between LIWC category scores and pain catastrophizing score.<br>2. Multiple regression analysis to identify which LIWC categories most contributed for the prediction of pain catastrophizing.<br>3. Regression analysis to identify LIWC category contributors but controlling for pain intensity and neuroticism. |
| [56] | 1. Extract 5 LIWC features: first-person pronouns, positive emotions, negative emotions, cognitive processes, biological processes.<br>2. Two annotators label emotional tone of each essay (classes: positive, negaive, mixed). | 1. Correlate LIWC features with psychological and physical health parameters (Hierarchical Linear Regression Analyses).<br>2. Correlate human-extracted features with psychological and physical health parameters. (Hierarchical Linear Regression Analyses).<br>3. Compare correlation models (LIWC vs. human extraction). |

Table A.10. Main lexico-semantic analysis methodology.

| Ref. | Feature extraction | Analysis |
|---|---|---|
| [35] | 1. Generate word frequency and keyword lists, and concordance plots with AntConc [2] (not described).<br>2. POS-tagging.<br>3. Filter out everything but nouns. | 1. Find relations between words that frequently co-occur using ManyEyes [3]. |
| [14] | 1. Define set of 24 terms related to pain and count their frequency.<br>2. Manual topic categorization of each tweet, where topics were pre-defined by the authors.<br>3. Manual gender categorization of each tweet, based on what the contributor displayed on the twitter bio, when available. | 1. Analysis of reach of each tweet, based on the number of impressions.<br>2. Topic analysis.<br>3. Sentiment analysis of each tweet based on Sentiment Dictionary that states the sentiment (-5 to +4) of more than 10,000 words. It was required that at least 2 words per tweet were recognized by the dictionary to assign a sentiment. |
| [15] | 1. Manually develop 3 vocabularies: body regions, types of healthcare providers, and alternative therapies.<br>2. Count frequency of each word in each vocabulary.<br>3. Topic modelling with LDA, k=1 according to coherence score. | 1. Wordcloud for (key)word frequency analysis.<br>2. Topic analysis. |
| [16] | 1. Extract adjectives from social media posts containing the word "pain". | 1. Word co-occurrence network analysis.<br>2. Qualitative analysis of a sample of social media posts, assigning topical dimensions to each.<br>3. Validated dimensions using W2V model and clustering (K-Means) on full dataset. |
| [20] | 1. Apply Leximancer [4] with 2 setups: automatically- and manually-seeded concept mappings, which is similar to topic modelling. The manually-seeded concepts were focused on pain management to skew the analysis to that direction. | 1. Study concept relation within the text and compare between groups (patients, general practitioners, and specialists). |
| [57] | 1. Select EHRs and blog posts that contain predefined terms related to pain, drugs, and nurse calls. | 1. Compare physician and patient perspective using correspondence analysis (not described). |
| [58] | 1. DLA on the documents – similar to topic modelling but finds the words that most correlate to given terms, in this case opioid medication (not described). | 1. Creation of word clouds from the DLA topics for qualitative analysis.<br>2. Topic analysis. |
| [36] | 1. Topic modelling with LDA - separate model for coaches and participants - k = 15 topics from qualitative gauge.<br>2. Transform each communication (set of documents from a single source – either parent-teen pair or coaches) into a topic vector, where +1 is added to the corresponding topic dimension that each message is labelled. | 1. Topic analysis.<br>2. Analysis of communication flow.<br>3. Communication clustering and analysis using K-Means (k=4). |
| [17] | 1. Thematic analysis on 4 sources of data: CRIS [5], MIMIC-III [59], Reddit, Twitter.<br>2. Extract pain-related words from 3 scientific publications.<br>3. Extract pain synonyms from 3 biomedical ontologies (UMLS, SNOMED-CT, ICD-10).<br>4. Using 8 different pre-trained word-embedding models, extract all words similar to "pain" according to a similarity cut-off point using the elbow method.<br>5. Lower-case and remove duplicates.<br>6. Add new terms to lexicon as suggested by clinicians. | 1. Validate and filter lexicon with clinicians. |
| [37] | 1. Define set of 78+51 pain descriptors (original MPQ [60] + thesaurus).<br>2. Using W2V, compute similarity score between each word in the sample with the pain descriptors.<br>3. Select the top-most similar words for each pain descriptor (maximum of 20). Manually curate new words for relevancy. | 1. Count the frequency of each pain descriptor and new words within the sample. Use counts of the original 78 MPQ pain descriptors as the minimum threshold for relevancy and discard all descriptors less frequent than that threshold.<br>2. Calculate intensity (score 0-1) of each pain descriptor using CrystalFeel [38], and order pain descriptors (within each category) according to intensity score.<br>3. Propose updated MPQ. |

---

[2] https://www.laurenceanthony.net/software/antconc/

[3] https://www.bewitched.com/manyeyes.html (tool not available since 2015)

[4] https://www.leximancer.com/

[5] https://www.maudsleybrc.nihr.ac.uk/facilities/clinical-record-interactive-search-cris/

Table A.11. Evaluation methods.

| Ref. | k-fold CV | Accuracy | $F_1$ | Precision/PPV | Recall/Sensitivity | Specificity | NPV | AUC | Qualitative | Other |
|---|---|---|---|---|---|---|---|---|---|---|
| [21] | | | | | ✓ | ✓ | | | | |
| [18] | ✓ 10 | ✓ | | ✓ | ✓ | ✓ | ✓ | | | |
| [2] | ✓ 10 | | ✓ | ✓ | ✓ | | | ✓ PRC, ROC | | |
| [3] | ✓ 5 | ✓ balanced | ✓ macro, micro, weighted | ✓ | ✓ | | | | | ✓ Mathew's correlation coefficient |
| [4] | ✓ 5, n | ✓ | ✓ weighted, micro | ✓ | ✓ | | | | | |
| [22] | ✓ n | | ✓ weighted | | | | | | | |
| [23] | | | | | | | | | ✓ | |
| [24] | ✓ 10 | | ✓ | ✓ | ✓ | | | ✓ | | |
| [25] | ✓ 10 | ✓ | ✓ | ✓ | ✓ | | | | | |
| [5] | | | ✓ | ✓ | ✓ | | | | | |
| [26] | | | ✓ | ✓ | ✓ | | | | | |
| [6] | ✓ 5 | ✓ | ✓ | ✓ | ✓ | ✓ | | ✓ PRC, ROC | | |
| [7] | ✓ 10 | | ✓ | ✓ | ✓ | | | | | |
| [27] | ✓ 2 | | | | | | | ✓ ROC | | |
| [28] | | ✓ | | | ✓ | ✓ | | | | ✓ correlation |
| [30] | | ✓ balanced | ✓ | ✓ | ✓ | | | ✓ ROC | | |
| [31] | ✓ 10 | ✓ | | | | | | | ✓ | ✓ perplexity, topic coherence |
| [8] | | | | | | | | | ✓ | ✓ Cohen's Kappa |
| [9] | | ✓ | ✓ | ✓ | ✓ | | | | | ✓ univariate and multivariate regression |
| [54] | | | | ✓ | ✓ | ✓ | ✓ | ✓ ROC | | |
| [55] | | | | | | | | | | ✓ correlation metrics |
| [56] | | | | | | | | | | ✓ beta unstandardized regression coefficient |
| [32] | | | ✓ | ✓ | ✓ | | | | | |
| [19] | ✓ 5 | | ✓ | ✓ | ✓ | | | ✓ ROC | | |
| [40] | | | | | | | | | | ✓ algorithm coverage vs. expert coverage |
| [10] | | | | | | | | | | ✓ inter-annotator agreement (accuracy and Cohen's Kappa) |
| [33] | | | ✓ | ✓ | ✓ | | | | | |

| Ref | | | | | | | | | | |
|------|---|---|---|---|---|---|---|---|---|---|
| [11] | | | | | | | | | ✓ | |
| [12] | ✓ 6 | | ✓ | ✓ | ✓ | | | | | ✓ inter-annotator agreement (Cohen's Kappa, Krippendorff's alpha) |
| [13] | | | | | | | | | ✓ | |
| [35] | | | | | | | | | ✓ | |
| [14] | | | | | | | | | | |
| [15] | | | | | | | | | | |
| [16] | | | | | | | | | ✓ | |
| [20] | | | | | | | | | ✓ | |
| [57] | | | | | | | | | ✓ | |
| [58] | | | | | | | | | | |
| [36] | | | | | | | | | ✓ manual verification of topic assignment | |
| [17] | | | | | | | | | ✓ | |
| [37] | | | | | | | | | ✓ | |

Table A.12. Primary outcome.

| Ref. | Primary outcome (question) | Primary outcome (conclusion) |
|---|---|---|
| [21] | Can a machine learning framework accurately classify patterns of LBP from EHRs? | High performance scores on a very small dataset. Requires larger data for a clear indicator of success. |
| [18] | What are clinical clues that distinguish TMD from TMD-mimicking conditions? | Mouth opening limitation and lack of mention of TMD joint or noise are more associated with mimicking conditions based on chief complaints. Cutoff mouth opening sizes of 12mm comfortable and 26.5mm maximum optimally distinguish between groups. |
| [2] | Can we automatically identify EHRs reporting acute LBP episodes? | Developed model is robust to the reduction of annotated training samples. Topic models can be used in an unsupervised setting. |
| [3] | How has back pain complaint on Twitter vary (in terms of frequency), from before to during the COVID-19 pandemic? | Percentage of back pain complaints in twitter increased 84% during COVID-19 pandemic in comparison to previous year. |
| [4] | Can machine learning algorithms accurately classify migraine and cluster headache patient narratives? | Developed model identifies differences in vocabulary for each class, which are successfully used for classification. These lexical characteristics are congruent with expert knowledge. |
| [22] | Can language features extracted from patient narratives estimate pain intensity? | Language features allow estimation of pain intensity with weighted $F_1$ of 0.60. Focus on specific words/themes correlates with specific pain intensities. |
| [23] | Can NLP techniques categorize EHRs based on osteoarthritis pain severity? | The approach shows potential for identifying patients with mild to severe osteoarthritis pain. Further development is needed. |
| [24] | Can we automatically detect EHRs that contain documentation of pain assessment? | Task is feasible with high performance scores. Requires further development to automatically assess PCQ indicators. |
| [25] | Can machine generated RegEx be used to improve classifications tasks of clinical text? | Machine-generated RegEx can be effectively used in clinical text classification. The RegEx-based classifier can be combined with other classifiers to improve classification performance. |
| [5] | Can an NLP pipeline trained on public EHRs data extract physician-reported pain from institutional EHRs? | Developed model successfully detects and quantifies reports of pain severity in EHRs. |
| [26] | Can EHRs be used to predict pain relevance and pain change in sickle cell disease patients? | DT and FFNN are promising models to predict pain relevance and pain change from EHRs in sickle cell disease patients. |
| [6] | Are NLP algorithms useful for identifying different attributes of pain interference (and fatigue) symptoms experienced by child and adolescent cancer survivors as compared to the judgment by PRO content experts? | As an alternative to using standard PRO surveys, collecting unstructured PROs via interviews or conversations during clinical encounters and applying NLP methods can facilitate PRO assessment in child and adolescent cancer survivors. |
| [7] | Can free-text patient questionnaire responses be automatically classified on a Likert scale for a closed-ended questionnaire? | Demonstration of the feasibility of processing open-ended patient questionnaire answers, in order to map to closed-ended questionnaires, in this case, Likert scales. |
| [27] | Can a deep learning model accurately identify risk factors for LBP in EHRs? | The application of BERT models on down-sampled annotated EHRs is useful in detecting risk factors suggesting an indication for imaging for patients with LBP. |
| [28] | Can language features identify placebo responders in chronic back pain? | Language features successfully identified placebo responders with 79% accuracy. |
| [30] | Can patient narratives prior to treatment be used to predict placebo and drug response? | Placebo response is predictable using NLP techniques and can be examined through the study of mental processes that are reflected onto the semantic content of patient narratives. |
| [31] | Can NLP techniques triage musculoskeletal patients based on referral letters? | Demonstration of the feasibility of automatically triaging knee or hip pain patients based on the contents of their referral letters. The latent topics modelled from the training data were shown to have relevant clinical interpretability. |
| [8] | Is increased attribution of chronic back pain to mind/brain processes associated with reductions in pain intensity? | Pain reprocessing therapy led to significant increases in mind- or brain-related attributed causes of pain and increases in mind-brain attributions were associated with reduced pain. |
| [9] | Can longitudinal NLP analysis of EHRs from metastatic prostate cancer patients be used to meaningfully depict the experience of their pain? | It is feasible to track longitudinal patterns of pain with text mining of free text from EHRs in a cohort of patients. Their model is also generalizable to other datasets, and it provides a number of phenotype-oriented observations useful for future research and for monitoring pain management and identifying novel cancer phenotypes. |
| [54] | Can NLP of EHRs be used to effectively assess risk of opioid-agreement violation? | Patients classified as high risk by the model were found to be three times more likely to violate the agreement than those classified with lower risk. |
| [55] | Are linguistic patterns associated with pain catastrophizing in patients with persistent musculoskeletal pain? | Pain catastrophizing is associated with a "linguistic fingerprint" that can be discerned from patients' natural word use. Patients who catastrophize exhibit a heightened focus on the self and the negative emotional aspects of their pain experience when writing about their life with chronic pain. |

| [56] | How do human raters and automated text analysis compare in predicting self-reported psychological and physical health? | The utility of automated text analysis over human raters depends on the individual characteristic being measured. Human ratings were better for predicting depression, but automated text analysis was sufficient for predicting pain catastrophizing and illness intrusiveness. |
|---|---|---|
| [32] | Can NLP efficiently and accurately extract pain information from EHRs? | The customized NLP model demonstrated good and successful performance in extracting granular pain information from EHRs. |
| [19] | Can an exemplar-based method accurately assign medical ontology classes to EHRs free-text? | The exemplar-based method achieved good precision and recall in assigning chronic pain ontology classes to unseen EHRs free-text. |
| [40] | How does NLP compare to medical experts in ICF factor extraction from EHRs? | Comparison between model outputs and medical expert labelling in terms of coverage reveals that the combination of both experts and model produces the best performance. |
| [10] | How is pain mentioned in mental health records? | Most pain mentions were relevant to patient's physical pain. Chest was most common anatomy. Chronic was most common character. |
| [33] | Can EHRs of the VHA be used to automatically extract PCQ indicators? | PCQ indicators can be reliably extracted from the VHA EHRs using NLP. |
| [11] | What are the patterns of pain care quality indicators documented by chiropractors across different visit types? | VHA chiropractors frequently document PCQ indicators, identifiable using NLP, with variability across different visit types. More total indicator classes were documented during consultation versus follow-up visits, with high co-occurrence of pain assessment indicators. |
| [12] | Can we accurately extract SDoH information from EHRS of chronic LBP patients? | The hybrid and machine learning models showed promising performance for extracting many SDoH entities, while rule-based cTAKES had lower performance. |
| [13] | What factors predict optimal versus suboptimal care pathway? | Lower body mass index, named disease/syndrome, taking pharmacologic substance predict optimal pathway. Single diagnostic procedure predicts suboptimal pathway. |
| [35] | Is text mining useful to characterize clinical distinctions and patient concerns of fibromyalgia in online patient-generated text? | Text mining has the potential for extracting keywords to confirm the clinical distinction of a certain disease and can help objectively understand the concerns of patients by generalizing their large number of subjective illness experiences. |
| [14] | Based on Twitter, which pain-related topics are discussed, which keywords are commonly used, what is the dissemination impact, and what are the user's genders? | Analysis highlighted prevalent keywords such as headache and migraine, with a higher engagement from females, and an overall negative sentiment. Retweets tended to convey more positive sentiment, notably in the cannabis and fibromyalgia categories. |
| [15] | What are the main discussion topics related to chronic pain, that patients discuss on social media? | Patients discuss body regions and compare treatments proposed by medical physicians versus other healthcare providers. |
| [16] | What can discussions of pain tell us about the social and communicative contexts of online endometriosis communities? | Sharing experiences of pain online provides validation and reinforces patient ownership and authority. |
| [20] | How do doctors and patients communicate about managing musculoskeletal disorders? | Doctors and patients emphasized different aspects - patients wanted to return to normal while doctors encouraged accepting a new normal. |
| [57] | What are the differences in perspectives on pain between nurses' observations and patients' experiences? | Observed differences between physicians and patients, namely on the early detection of side effects. |
| [58] | Are Yelp reviews useful for understanding the patients' and caregivers' experiences related with pain management and opioids? | Yelp reviews offer insights into pain management and opioid use that are not assessed by traditional surveys. |
| [36] | What are the patterns of engagement during an internet-delivered cognitive behavioral therapy intervention for youth with chronic pain? | Identified 4 clusters of participants based on patterns of engagement - Assignment-Focused, Short Message Histories, Pain-Focused, and Activity-Focused |
| [17] | What are key terms to describe pain? | Proposed lexicon consists of 382 terms, derived from 3 sources. Validated by two clinicians and compared against existing ontologies. |
| [37] | What are relevant contemporary English pain descriptors based on social media data? | Suggest removing 11 descriptors from the original MPQ, adding 13 new descriptors to existing categories, and adding a new category (psychological) with 9 new descriptors. Modified MPQ requires validity analysis. |

**Acronyms**

| | |
|---|---|
| AUC | Area Under the Curve |
| BOB | Bag-of-Bigrams |
| BOW | Bag-of-Words |
| CNN | Convolutional Neural Network |
| CUI | Concept Unique Identifier |
| CV | Cross Validation |
| DLA | Differential Language Analysis |
| DT | Decision Tree |
| EHR | Electronic Health Record |
| FFNN | Feed Forward Neural Network |
| ICD | International Classification of Diseases |
| ICF | International Classification of Functioning, Disability, and Health |
| kNN | k-Nearest Neighbors |
| KOOS | Knee injury and Osteoarthritis Outcome Score |
| LBP | Low back pain |
| LDA | Latent Dirichlet Allocation |
| LR | Logistic Regression |
| LSA | Latent Semantic Analysis |
| MeSH | Medical Subject Headings |
| MPQ | McGill Pain Questionnaire |
| NB | Naïve Bayes |
| NER | Named Entity Recognition |
| NLP | Natural Language Processing |
| NPV | Negative Predictive Value |
| PCQ | Pain Care Quality |
| POS | Part-of-Speech |
| PPV | Predictive Positive Value |
| PRC | Precision-Recall Curve |
| PRO | Patient Reported Outcome |
| PROM | Patient Reported Outcome Measurement |
| RegEx | Regular Expression |
| RF | Random Forest |
| ROC | Receiver Operating Characteristic |
| SDoH | Social Determinants of Health |
| SNOMED-CT | Systematized Nomenclature of Medicine – Clinical Terms |
| SVM | Support Vector Machine |
| TF-IDF | Term Frequency – Inverse Document Frequency |
| TMD | Temporomandibular |
| UMLS | Unified Medical Language System |
| VDP | Verbally Declared Pain |
| VHA | Veteran's Health Administration |
| W2V | Word2Vec |